\documentclass[10pt,twocolumn,letterpaper]{article}

\usepackage[pagenumbers]{cvpr}

\usepackage{colortbl}
\usepackage{booktabs}

\usepackage{microtype}
\newcommand{\UCFOfficialMAE}{0.115}
\newcommand{\UCFOBO}{87.62}

\usepackage{graphicx}

\extrafloats{100}

\definecolor{cvprblue}{rgb}{0.21,0.49,0.74}
\usepackage[pagebackref,breaklinks,colorlinks,allcolors=cvprblue]{hyperref}

\def\confName{CVPR}
\def\confYear{2027}

\title{DiffRAC: Repetitive Action Counting with Video Diffusion Motion Priors }
\author{%
Junyuan Xiao$^{1}$\thanks{Equal contribution.} \quad Guangmo Yi$^{1}$\footnotemark[1] \quad Jingqi Tian$^{1}$ \quad Cheng Chen$^{2}$\\
Lionel Z. WANG$^{3}$ \quad Wenming Yang$^{1}$\thanks{Corresponding author.} \quad Xinjie Li$^{4}$\\[0.4em]
{\normalsize $^{1}$Tsinghua University \quad $^{2}$Xiaohongshu}\\
{\normalsize $^{3}$Hong Kong Polytechnic University \quad $^{4}$Pennsylvania State University}\\[0.2em]
{\small\ttfamily xiao-jy24@mails.tsinghua.edu.cn}%
}
\hypersetup{%
  pdftitle={DiffRAC: Repetitive Action Counting with Video Diffusion Motion Priors},
  pdfauthor={Junyuan Xiao, Guangmo Yi, Jingqi Tian, Cheng Chen, Lionel Z. WANG, Wenming Yang, Xinjie Li}%
}

\begin{document}
\maketitle

\begin{abstract}
Repetitive action counting (RAC) requires identifying complete cycles in continuous motion, but visual similarity is an unreliable cue for this: a pose can recur within a single cycle, and pauses or static backgrounds produce high similarity without any repetition. We argue that video diffusion models, which must capture how content evolves across frames in order to generate coherent motion, provide a better source of cross-frame relations. We present DiffRAC, which uses a Wan video diffusion transformer, adapted with LoRA and then frozen, as the motion encoder of a counting model. For each video window, DiffRAC reads the query--key attention affinities between time steps of the input clip from several denoiser layers and averages them along temporal lags. Regularly spaced lag responses, summarized by learned lag features, a periodogram, and period templates, yield a posterior over the cycle length. Conditioned on this posterior, a query-based decoder predicts whether the window is repeating (activity) and how fast it repeats (frequency); their product is a local counting rate. Rates from overlapping windows are averaged on the original timeline by coverage normalization, integrated into a count, and calibrated, and an optional evidence-gated refinement step can further update the count. DiffRAC reaches 74.3\% off-by-one accuracy (OBO) on RepCount-A and 87.6\% on UCFRep. Trained only on RepCount-A, it transfers to UCFRep and Countix with 82.4\% and 73.9\% OBO.
\end{abstract}

\section{Introduction}
\label{sec:intro}

\suppressfloats[t]
\begin{figure}[t]
\centering
\includegraphics[width=0.9\columnwidth]{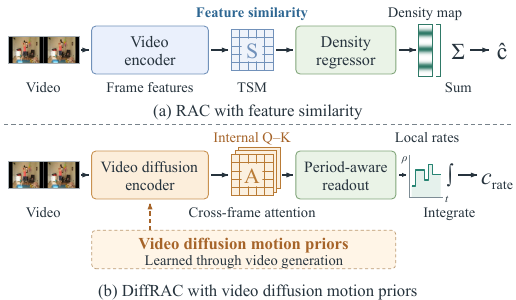}
\caption{\textbf{Two RAC paradigms.} (a) Representative methods regress density maps from feature self-similarity and sum them into counts. (b) DiffRAC uses video diffusion motion priors via cross-frame attention and period-aware readout, integrating local rates into $c_{\mathrm{rate}}$ before boundary correction and count calibration.}
\label{fig:rac_paradigms}
\end{figure}

Recent advances in spatiotemporal modeling and pretrained video representations~\cite{carreira2017quo,feichtenhofer2019slowfast,videoswin,videomae} have provided a foundation for fine-grained action analysis.
Repetitive action counting (RAC) is an important task in this area, aiming to estimate how many times an action is completed in a video~\cite{repnet,transrac}.
It supports applications in fitness monitoring, sports analysis, and rehabilitation assessment~\cite{countllm}, converting continuous motion into counts that quantify exercise volume and inform training assessment.

The main difficulty of RAC is that its usual cue, similarity between frames, is ambiguous.
Cycle lengths vary across actions and within a video as people speed up, slow down, or pause~\cite{zhang2020context,wang2024lmrl}; a single cycle can contain similar moments, such as the descending and ascending phases of a squat; and backgrounds and camera motion distort inter-frame relations~\cite{repnet,transrac,li2024mfl}.
Prior methods derive period cues from optical flow~\cite{runia2018real} or from the similarity of learned features, as in RepNet's temporal self-similarity matrix (TSM)~\cite{repnet} and TransRAC's multi-scale correlations~\cite{transrac} (Fig.~\ref{fig:rac_paradigms}(a)), with later work adding cycle alignment~\cite{luo2024rethinking} or language priors~\cite{countllm}.
Their period evidence is bounded by the representation: when features do not separate motion phases, high similarity need not indicate a repeated cycle~\cite{luo2024rethinking,wang2024lmrl}.
This raises the question of which pretrained representation encodes motion in a way that makes complete cycles identifiable.

Video diffusion models are a natural candidate.
To generate videos with coherent motion~\cite{ho2022video,blattmann2023align,wan}, their denoisers must model how each region changes from frame to frame, and this information is exposed in their attention.
Motion-transfer methods reuse cross-frame attention to reproduce the motion of a reference video~\cite{ling2025motionclone,pondaven2025ditflow}, and DiffTrack~\cite{difftrack} shows that queries and keys (Q--K) from selected layers of a video diffusion transformer match points across real frames.
Tracking, however, follows the same point as it moves, whereas counting asks when the action returns to a corresponding phase.
We hypothesize that the denoiser assigns high affinity to corresponding phases of successive cycles, so that the time intervals between high-affinity pairs reveal the period more reliably than generic feature similarity does.
Sec.~\ref{sec:repcount_ablation} tests this hypothesis by comparing Q--K affinities with hidden-feature similarity from the same backbone under an identical readout.

Building on this hypothesis, we propose \textbf{DiffRAC}, a repetitive action counting framework that uses a pretrained Wan video diffusion transformer~\cite{wan22} as its motion encoder (Fig.~\ref{fig:rac_paradigms}(b)).
DiffRAC is trained in two stages. We first adapt Wan to counting videos with LoRA~\cite{lora} under its original flow-matching objective and then freeze it; we then train a lightweight counting readout on affinities cached from the frozen encoder.
At inference, the video is split into overlapping windows, and each window passes through four steps.
(i)~\emph{Motion encoding.} One denoising forward pass on the noised latent of the window exposes the self-attention between its time steps, which we pool over space into temporal Q--K affinity maps from several layers and heads.
(ii)~\emph{Period evidence.} A cycle of length $p$ produces bands parallel to the diagonal of these maps at offsets $p, 2p, \ldots$. Averaging each offset diagonal gives a lag profile, from which learned lag features, a fixed periodogram, and period-template scores jointly estimate a posterior over $p$.
(iii)~\emph{Rate decoding.} Following RepNet's distinction between periodicity and period length~\cite{repnet}, a period-conditioned query decoder predicts how strongly the window is repeating (activity) and how fast (frequency); their product is a local rate in repetitions per frame.
(iv)~\emph{Counting on the original timeline.} Because windows overlap and can be sampled at different strides, we map each rate back to the original frames, average the overlapping rates within each time bin, integrate the result, and calibrate the integral into a count.

Because the readout exposes interpretable intermediate quantities, namely period posteriors, activities, and rates, its count can be checked against other evidence.
DiffRAC therefore includes an evidence-gated refinement step. It proposes an alternative count, obtained from a second readout, from additional affinity-based heads, or from auxiliary pose and action cues, and accepts the proposal only when the corresponding agreement criterion is met; otherwise the initial count is kept.

Our main contributions are as follows:
\begin{enumerate}[leftmargin=2em,label=\arabic*.]
\item We introduce DiffRAC, to our knowledge the first repetitive action counting framework that uses a pretrained video diffusion model directly as its motion encoder.
\item We propose a period-aware readout that converts multi-layer Q--K affinities into a posterior over cycle length by aggregating them along temporal lags, and decodes each window's counting rate as the product of repetition activity and frequency.
\item We introduce coverage-normalized integration, which counts overlapping windows sampled at different strides on a single original timeline, and combine it with evidence-gated count refinement.
\item DiffRAC reaches 74.34\% OBO on RepCount-A, the highest among the methods in Tab.~\ref{tab:reported_results}, and \UCFOBO\% OBO on UCFRep. Trained only on RepCount-A, it exceeds the OBO of CountLLM~\cite{countllm} on both UCFRep and Countix (Tab.~\ref{tab:cross_dataset}).
\end{enumerate}
\begin{figure*}[t]
\centering
\includegraphics[width=\textwidth]{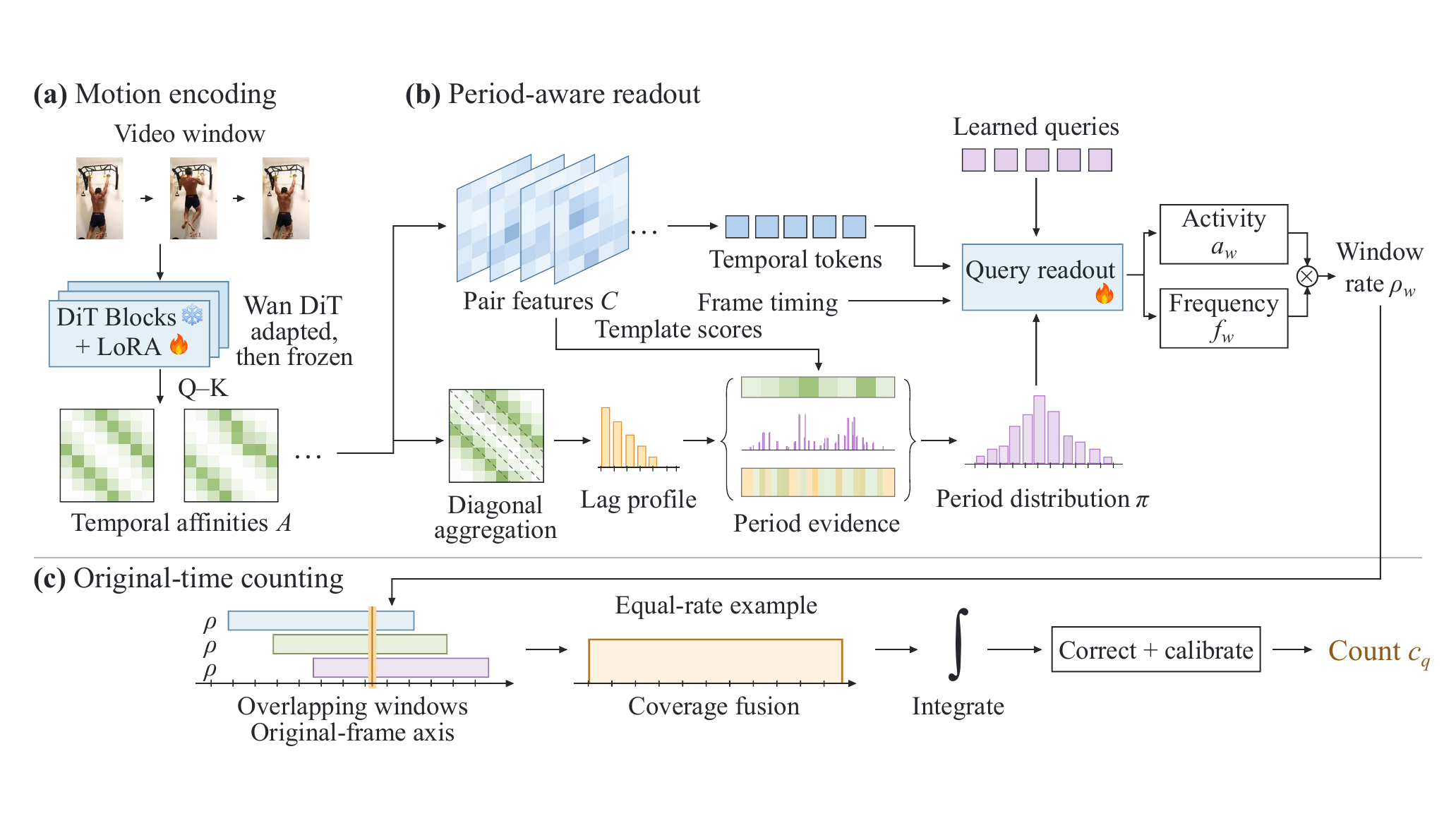}
\vspace{-10pt}
\caption{\textbf{Overview of the DiffRAC query readout.} (a)~Wan, adapted with LoRA in Stage~1 and then frozen, extracts multi-layer temporal Q--K affinities $A$ from a video window (Sec.~\ref{sec:representation}). (b)~The readout, trained in Stage~2, combines lag, spectral, and template evidence into a period distribution $\pi$ (Sec.~\ref{sec:period_model}); learned queries read temporal tokens pooled from pair features $C$ together with $\pi$ and the frame timing, and predict activity $a_w$ and frequency $f_w$, whose product is the window rate $\rho_w$ (Sec.~\ref{sec:decoding}). (c)~Overlapping rates are averaged per time bin, integrated, and calibrated into the count $c_q$ (Sec.~\ref{sec:aggregation}).}
\label{fig:framework}
\end{figure*}

\begin{figure*}[t]
\centering
\includegraphics[width=\textwidth]{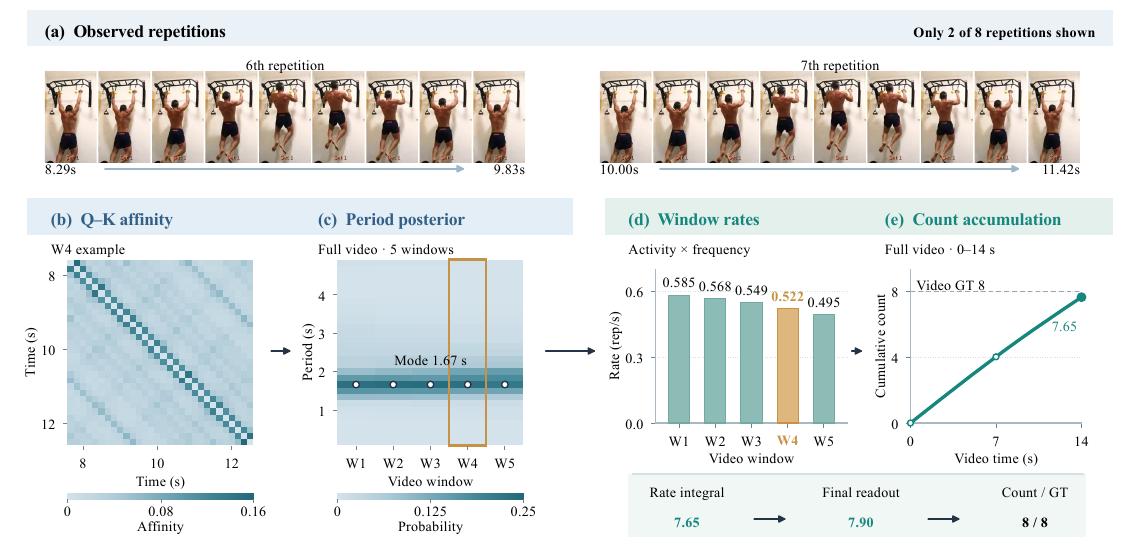}
\caption{\textbf{From recurrence to count on a 14-second RepCount-A test clip with eight pull-ups}, using one trained query readout. (a)~Frames of repetitions 6 and 7. (b)~Channel-averaged Q--K affinities of window W4. (c)~Period posteriors of the five windows; circles mark modes, gold marks W4. (d)~Window rates $a_wf_w$ in repetitions per second. (e)~Coverage-normalized rate integral and calibrated prediction; the dashed line is the annotated total.}
\label{fig:mechanism_real}
\end{figure*}

\section{Related Work}

\paragraph{Repetitive action counting.}
Existing methods differ in where they obtain the cue that two moments correspond across cycles. Motion-signal methods analyze optical flow with wavelets~\cite{runia2018real} or use flow-guided foreground features~\cite{li2024mfl}. Similarity-based RGB methods compare encoder features over time: RepNet~\cite{repnet} predicts periodicity and cycle length from a temporal self-similarity matrix, TransRAC~\cite{transrac} regresses density from multi-scale correlations, and follow-ups make the similarity context-aware~\cite{zhang2020context}, localization-aware~\cite{wang2024lmrl}, or robust to irregular repetitions~\cite{ivacp2l}, align phases with cycle annotations~\cite{luo2024rethinking}, or match exemplars~\cite{escount}; CountLLM~\cite{countllm} guides a language model with periodicity queries and text instructions. Skeleton-based methods count from estimated poses using joint-wise self-similarity~\cite{jtsps}, salient poses~\cite{poserac}, geometric or graph-based body features~\cite{gmfl,bigc}, or sparse attention on reconstructed 3D motion~\cite{mocount}. Apart from flow-based methods, their encoders are trained for recognition, pose, or language; DiffRAC instead reads the attention of a video generator and conditions a per-window rate on a period distribution.

\paragraph{Motion priors in video diffusion models.}
Video diffusion pretraining improves action recognition and point tracking over image diffusion pretraining~\cite{velez2025imagetovideo}. Cross-frame attention transfers motion in generation and editing~\cite{ling2025motionclone,pondaven2025ditflow,gu2024videoswap}, and Q--K similarities~\cite{difftrack} and adapted diffusion features~\cite{son2026ditracker} support point tracking on real videos. These works follow the \emph{same content} across frames; DiffRAC uses the \emph{temporal separation} between frames that attend to each other, which reveals when an action returns to a corresponding phase.

\paragraph{Video generation models for video understanding.}
GenDSR~\cite{yang2026gendsr} transfers generative features to dynamic spatial reasoning, and MotionEnhancer~\cite{xu2026motionenhancer} uses diffusion attention as auxiliary supervision for a vision-language model and evaluates counting as video question answering. DiffRAC instead uses the adapted denoiser itself as its encoder at inference and counts through explicit, inspectable period evidence and local rates.

\section{Method}
\label{sec:method}

\subsection{Problem Formulation and Overview}
\label{sec:problem}
\label{sec:overview}

Given an $N$-frame RGB video $V$ and conditioning $u$, such as a text prompt $u_{\rm text}$ or action metadata, repetitive action counting estimates a nonnegative count $\hat c=f_\Theta(V,u)$ of complete cycles; no ground-truth count is used at inference.
DiffRAC splits $V$ into $W$ overlapping windows $V_w$ (Fig.~\ref{fig:framework}). The frozen encoder maps each window to temporal affinity maps $A_w$ (Sec.~\ref{sec:representation}). From $A_w$, the readout estimates \emph{which period} the window supports (Sec.~\ref{sec:period_model}), decides \emph{how much} the window counts as a rate $\rho_w$ (Sec.~\ref{sec:decoding}), and integrates the rates on the original timeline into a count $c_q$ (Sec.~\ref{sec:aggregation}). We call these three steps the \emph{query readout}. Optional \emph{count refinement} updates $c_q$ only when a gate accepts the change (Sec.~\ref{sec:refinement}). The encoder (Stage~1) and the readout (Stage~2) are trained separately (Sec.~\ref{sec:optimization}).

\subsection{Motion Encoding with Video Diffusion Priors}
\label{sec:representation}
\paragraph{Diffusion encoder.}
To generate coherent videos, a diffusion denoiser must model how content evolves across frames~\cite{ho2022video,wan}, and selected layers of its self-attention retain inter-frame correspondences~\cite{ling2025motionclone,pondaven2025ditflow,difftrack}. DiffRAC uses the \emph{temporal separation} of these correspondences as its counting cue. We use Wan2.2 TI2V-5B~\cite{wan,wan22} adapted with LoRA~\cite{lora} (Sec.~\ref{sec:optimization}). For an encoded window $z_0$, the denoiser receives $z_\sigma=(1-\sigma)z_0+\sigma\varepsilon$ with Gaussian noise $\varepsilon$ and noise level $\sigma\in[0,1]$. Because it denoises a latent of the input clip, its attention is conditioned on the observed sequence, and every affinity maps to a pair of known input times. One forward pass per window suffices, and all weights stay frozen.

\paragraph{Cross-frame query--key affinities.}
The VAE maps the 121 sampled frames of a window to $T_0=31$ latent nodes. Under the video-to-video schedule, we capture queries and keys after Q--K normalization and rotary embedding. For layer--head channel $h\in\{1,\ldots,H\}$, with $S$ spatial tokens per node and head width $d_h$, we compute attention over all video keys and average over spatial pairs:
\begin{align}
B_h &= \operatorname{softmax}_{\rm key}
       (Q_h K_h^{\mathsf T}/\sqrt{d_h}),\label{eq:attention}\\
\widetilde A_h[i,j] &=
  \frac{1}{S^2}\sum_{m,m'=1}^{S}B_h[(i,m),(j,m')].
\label{eq:pool}
\end{align}
$\widetilde A_h[i,j]$ measures how strongly the denoiser relates times $i$ and $j$. Unlike hidden-feature self-similarity, \ie, cosine similarity of pooled hidden features (Sec.~\ref{sec:repcount_ablation}), it reflects the relations that the attention operation actually uses, and separate maps per head let the readout select channels that respond to recurrence.

Each map is symmetrized, its diagonal is removed, and its rows are normalized. We drop the first latent node, which encodes only the first frame in Wan's causal VAE, %
and renormalize the remaining $T=30$ nodes. All 24 heads of layers 7, 14, 21, and 27 give $A_w\in\mathbb R^{H\times T\times T}$ with $H=96$. Because similar poses within a cycle, pauses, and static background also produce high affinities, the readout relies on intervals supported by many node pairs and channels rather than on individual entries.

\subsection{From Diffusion Affinities to Period Evidence}
\label{sec:period_model}
This subsection and the next form the core of the query readout. Here we estimate \emph{which recurrence interval} a window supports; Sec.~\ref{sec:decoding} then uses this estimate to decide \emph{how many} repetitions the window contributes. Motion with period $p$ produces bands parallel to the diagonal at offsets $p,2p,\ldots$ (Fig.~\ref{fig:mechanism_real}(b)), which we summarize with three complementary types of evidence, namely lag, spectral, and template evidence, and combine into a posterior over periods. In parallel, a $1{\times}1$ projection and residual convolutions map $A_w$ to pair features $C\in\mathbb R^{D\times T\times T}$, which keep \emph{where} in the window the relations occur and are reused by the decoder. We omit $w$ below.

\paragraph{Lag evidence.}
For each channel, we average each offset diagonal:
\begin{equation}
r_h[\tau]=\frac{1}{T-\tau}\sum_{i=1}^{T-\tau}A_h[i,i+\tau],
\quad \tau=1,\ldots,T-1.
\label{eq:lag}
\end{equation}
$r_h[\tau]$ measures how much channel $h$ supports a recurrence after $\tau$ latent-node intervals. Averaging over starting positions turns pairwise relations into interval evidence that no longer depends on where the cycle starts. Dilated 1D convolutions over lags, followed by lag-axis averaging, give a learned lag descriptor $\boldsymbol\ell$. The mean lag response alone is not sufficient: a short lag can have high mean affinity simply because adjacent frames change little, whereas a full cycle of eight latent intervals produces peaks at lags 8, 16, and 24. The \emph{regular spacing} of these peaks, captured by the next two terms, is what separates a complete cycle from slow motion or a partial cycle.

\paragraph{Spectral evidence.}
With the channel-averaged profile $\bar r[\tau]=H^{-1}\sum_h r_h[\tau]$ and its mean $\mu_r$ over lags, a fixed periodogram over candidate periods $\mathcal P=\{1,\ldots,T-1\}$ is
\begin{equation}
P[p]=\left|\sum_{\tau=1}^{T-1}
(\bar r[\tau]-\mu_r)e^{-2\pi\mathrm i\tau/p}\right|^2.
\label{eq:periodogram}
\end{equation}
Standardizing $\log(1+P)$ over $\mathcal P$ gives the spectral evidence $\boldsymbol\phi$. Since $P[1]=0$, $\boldsymbol\phi$ discriminates only among $p\ge2$; it also responds to harmonics.

\paragraph{Template evidence.}
The periodogram operates on a fixed, channel-averaged profile. To score periods on the \emph{learned} pair features, we project $C$ to a score map $U\in\mathbb R^{T\times T}$ and compute $\psi[p]=\langle U,G_p\rangle$, where $G_p$ places Gaussian bands at multiples of $p$ (Appendix~\ref{sec:arxiv_period}). The template score aggregates node pairs whose separation is consistent with $p$ over the whole window, and therefore retains the spatial arrangement of the bands that the lag profile discards. The three types of evidence are combined into the period posterior
\begin{equation}
\boldsymbol\pi=\operatorname{softmax}\bigl(\operatorname{MLP}([\boldsymbol\ell;\boldsymbol\phi;\boldsymbol\psi])\bigr).
\label{eq:posterior}
\end{equation}

\paragraph{Node features.}
Because $\boldsymbol\pi$ pools over starting positions, it describes the whole window and can keep several modes under partial cycles, pauses, or harmonics. For local evidence, we average the row- and column-pooled features of $C$ at each node $i$ into $\mathbf x_i\in\mathbb R^{D}$.

\subsection{Window Rate Decoding}
\label{sec:decoding}
\label{sec:rate}
The decoder decides whether the window actually repeats at a supported period, and how fast. We embed $\boldsymbol\pi$, the original-frame position, and validity into each $\mathbf x_i$, and $J=32$ learned queries cross-attend to these tokens~\cite{countllm}. Their outputs, $\boldsymbol\ell$, $\boldsymbol\pi$, and the window geometry form $\mathbf z_w$. The geometry includes the effective stride $s_w$ in original frames per latent node, since the same lag implies different frequencies at different strides. Following RepNet~\cite{repnet}, we separate activity from frequency:
\begin{equation}
\begin{aligned}
a_w &= \operatorname{sigmoid}(\operatorname{MLP}_a(\mathbf z_w)),\quad
f_w = \operatorname{softplus}(\operatorname{MLP}_f(\mathbf z_w)),\\
\rho_w &= a_w f_w,\qquad \hat c_w=\rho_w\,|O_w|,
\end{aligned}
\label{eq:rate}
\end{equation}
where $O_w$ is the original-frame interval covered by the window's valid samples and $\hat c_w$ its expected count. The rate $\rho_w$ is in cycles per original frame and is constant within a window. The frequency $f_w$ is predicted from $\mathbf z_w$, which carries the full posterior rather than only its mode, so it can use the node features to choose among harmonics. The activity $a_w$ lets a window contribute little even when its posterior has a clear peak, \eg, during a pause between sets or before the first repetition starts. Activity, rate, and period each receive their own supervision (Sec.~\ref{sec:optimization}).

\subsection{Counting on the Original Timeline}
\label{sec:aggregation}
\label{sec:timeline}
\paragraph{Coverage-normalized integration.}
Summing $\hat c_w$ over overlapping windows would count shared frames repeatedly. We instead partition the original timeline into $B$ bins $I_b$ of width $\Delta=N/B$. Window $w$ covers $O_w$, from its first valid frame index to its last plus one, and overlaps bin $b$ by $\omega_{wb}=|O_w\cap I_b|$. With valid-window flags $v_w$,
\begin{equation}
\begin{aligned}
\hat\rho_b &=
\frac{\sum_w v_w\omega_{wb}\rho_w}{\sum_w v_w\omega_{wb}},
\qquad d_b=\Delta\hat\rho_b,\\
c_{\rm rate}&=\sum_b d_b .
\end{aligned}
\label{eq:fusion}
\end{equation}
Each bin is thus integrated once: windows that predict the same rate in a shared bin leave $d_b$ unchanged. Uncovered bins use the duration-weighted mean valid-window rate, or zero if no window is valid.

\paragraph{Boundary correction and count calibration.}
The integral $c_{\rm rate}$ is a continuous quantity, whereas the target is an integer number of complete cycles. Two small corrections close this gap. First, a boundary head predicts left and right boundary scores $\beta_w^{\rm L},\beta_w^{\rm R}\in(0,1)$, each scaled by the window activity, and adds a term for adjacent windows, $c_{\rm int}=c_{\rm rate}+\gamma\sum_{w} v_wv_{w+1}\beta_w^{\rm R}\beta_{w+1}^{\rm L}$, with learned $\gamma\in(0,1)$. Second, a bounded scalar calibration adjusts $c_{\rm int}$ in log space,
\begin{equation}
c_{q}=\max\bigl\{0,\exp[\log(1+c_{\rm int})+\delta]-1\bigr\},\quad
\delta=1.5\tanh f_{\rm cal}(\mathbf g),
\label{eq:count}
\end{equation}
where $f_{\rm cal}$ is a small MLP on video-level summaries $\mathbf g$: $\log(1+c_{\rm int})$, $\log(1+N)$, the mean activity, the mean period-implied and window counts, and the number of valid windows. The bound $|\delta|\le1.5$ limits the calibration to a multiplicative change of at most $e^{1.5}$ in $1+c_{\rm int}$, so the count remains anchored to the rate integral.

\begin{table*}[!t]
\caption{\textbf{Repetitive action counting on three benchmarks.} OBO is in percent. \textbf{Bold} and \underline{underline} mark the best and second-best value in each column. Published results follow the metric definitions and evaluation protocols of their sources. $^\dagger$DiffRAC on UCFRep uses action labels, annotated temporal crops, and two fixed per-video routing assignments (Sec.~\ref{sec:refinement}).}
\label{tab:reported_results}
\centering
\small
\setlength{\tabcolsep}{4pt}
\renewcommand{\arraystretch}{0.93}
\begin{tabular*}{\textwidth}{@{\extracolsep{\fill}}lllrrrrrr@{}}
\toprule
\textbf{Method} & \textbf{Publication} & \textbf{Family} & \multicolumn{2}{c}{RepCount-A} & \multicolumn{2}{c}{Countix} & \multicolumn{2}{c}{UCFRep} \\
\cmidrule(lr){4-5}\cmidrule(lr){6-7}\cmidrule(lr){8-9}
 & & & nMAE $\downarrow$ & OBO $\uparrow$ & nMAE $\downarrow$ & OBO $\uparrow$ & nMAE $\downarrow$ & OBO $\uparrow$ \\
\midrule
\multicolumn{9}{l}{\textit{Published results (source-specific protocols)}} \\
RepNet~\cite{repnet,repnet_eval} & CVPR 2020 & RGB & 0.331 & 53.29 & 0.364 & 69.66 & 0.209 & 73.33 \\
TransRAC~\cite{transrac} & CVPR 2022 & RGB & 0.443 & 29.13 & -- & -- & -- & -- \\
ESCounts~\cite{escount} & ACCV 2024 & RGB & 0.213 & 56.30 & 0.276 & 67.30 & 0.216 & 70.40 \\
IVAC-P2L~\cite{ivacp2l} & TMM 2025 & RGB & 0.402 & 34.44 & 0.507 & 43.09 & 0.503 & 42.00 \\
\addlinespace[2pt]
JTSPS-Net~\cite{jtsps} & TCSVT 2024 & Skeleton & 0.302 & 47.60 & -- & -- & 0.357 & 57.50 \\
GMFL-Net~\cite{gmfl} & arXiv 2024 & Skeleton & 0.216 & 58.60 & 0.269 & 59.40 & 0.259 & 65.00 \\
PoseRAC~\cite{poserac} & ICONIP 2024 & Skeleton & 0.226 & 57.00 & 0.305 & 53.00 & 0.146 & 80.30 \\
BIGC-Net~\cite{bigc} & EAAI 2025 & Skeleton & 0.215 & 57.90 & 0.268 & 60.60 & 0.287 & 59.80 \\
MoCount~\cite{mocount} & ACM MM 2025 & Skeleton & 0.174 & \underline{64.90} & -- & -- & 0.122 & 80.52 \\
\addlinespace[2pt]
CountLLM~\cite{countllm} & CVPR 2025 & LLM-based & \underline{0.162} & 63.90 & \underline{0.215} & \underline{80.10} & \underline{0.116} & \underline{83.90} \\
\midrule
DiffRAC (Ours) & --- & Diffusion-based & \textbf{0.143} & \textbf{74.34} & \textbf{0.166} & \textbf{83.64} & \textbf{\UCFOfficialMAE}$^\dagger$ & \textbf{\UCFOBO}$^\dagger$ \\
\bottomrule
\end{tabular*}
\end{table*}

\subsection{Evidence-Gated Count Refinement}
\label{sec:refinement}
Refinement treats $c_q$, or an estimate built from several readouts, as a reference and accepts a candidate update only when its gate criterion is met. We use one rule per benchmark:
\begin{itemize}[leftmargin=1.2em,itemsep=1pt,topsep=2pt]
\item \emph{Readout fusion} (RepCount-A). A low-count gate zeroes the count of readout R1 under weak repetition evidence; the count of a fine-tuned second readout R2 is accepted if it lies within 1.5 of the gated R1 count.
\item \emph{Affinity ensemble refinement} (UCFRep). The reference is the mode of 25 Wan-affinity readouts. Two density--rate heads with text-derived affinities and action-conditioned calibration propose replacements under action-specific rules; two action routes use fixed per-video assignments.
\item \emph{Teacher-local reranking} (Countix). The reference is a fixed \emph{teacher}, an ensemble of 14 self-similarity, VideoMAE-based, and DINO-based count experts, some using Wan lag features. A ranker scores integers within $\pm3$ of the teacher count from pose recurrence, action probabilities, and count--rate priors, and a learned gate accepts the best candidate only if it predicts a net benefit.
\end{itemize}
Pose closure and camera-compensated point tracks are used in the cross-dataset analyses.

\subsection{Two-Stage Training}
\label{sec:optimization}

\paragraph{Stage 1: diffusion adaptation.}
We train rank-32 LoRA on the attention and feed-forward layers of all 30 Wan blocks with Wan's flow-matching objective~\cite{fm}, regressing $\varepsilon-z_0$ from $z_\sigma$, on counting videos. Temporally shuffled negatives, which keep frame content but break motion order, balance each training set. %
The pretrained weights, VAE, and text encoder stay frozen.
Stage-1 adaptation captions include action and count descriptions. During affinity extraction for the RepCount-A query readouts and the UCFRep correction heads, we use a fixed instruction template whose sole video-specific field is the dataset-provided action name (Appendix~\ref{sec:arxiv_extraction}). The prompt contains no reference count or annotation-derived window answer.

\paragraph{Stage 2: readout learning.}
We cache $A_w$ for all training windows once and train the query readout with
\begin{equation}
\begin{aligned}
\mathcal L_{\rm read}
  &=\mathcal L_{\rm count}+\alpha_d\mathcal L_{\rm density}
    +\alpha_w\mathcal L_{\rm window}\\
  &\quad+\alpha_p\mathcal L_{\rm period}+\alpha_{\rm aug}\mathcal L_{\rm aug}.
\end{aligned}
\label{eq:loss}
\end{equation}
All targets derive from annotated cycle boundaries. $\mathcal L_{\rm count}$ applies Smooth-L1 to normalized, absolute, and log counts. Each cycle contributes a unit-mass Gaussian to a reference density on the $B$ bins~\cite{transrac}, which supervises the rescaled $d_b$; integrating it over $O_w$ gives the window count, rate, and activity targets of $\mathcal L_{\rm window}$. $\mathcal L_{\rm period}$ is a KL divergence between $\boldsymbol\pi_w$ and a Gaussian on $\mathcal P$ centered at the video-average period $N/(c\,s_w)$, on valid windows of videos with $c>0$. $\mathcal L_{\rm aug}$ assigns zero count and density to inputs with jointly permuted affinity rows and columns, which keep pair content but destroy lag structure. Appendix~\ref{sec:arxiv_readout_training} lists the weights.

\raggedbottom
\section{Experiments}
\label{sec:exp}

\subsection{Experimental Setup}
\label{sec:configs}

\paragraph{Datasets.}
We evaluate on RepCount-A~\cite{transrac} (758/131/152 training/validation/test videos with cycle boundaries), UCFRep~\cite{zhang2020context} (421 official training videos; with no test split, we report on the 105 validation videos), and Countix~\cite{repnet} (2,262 test annotations after the ESCounts filter~\cite{escount}: positive count and more than 40 frames).

\paragraph{Evaluated configurations.}
All configurations share the encoder and query readout and differ only in the refinement rule (Sec.~\ref{sec:refinement}): readout fusion on RepCount-A, affinity ensemble refinement on UCFRep, and teacher-local reranking on Countix. For transfer (Sec.~\ref{sec:cross_dataset}), all components are trained on RepCount-A only.

\paragraph{Evaluation metrics.}
We report off-by-one accuracy (OBO) and normalized mean absolute error (nMAE) under the ESCounts convention~\cite{escount}, with reference count $c_j$ and nearest-integer prediction $\hat c_j$ (ties to even):
\begin{equation}
\begin{aligned}\mathrm{OBO}&=\frac1n\sum_{j=1}^{n}\mathbf1(|\hat c_j-c_j|\leq1),\\\mathrm{nMAE}&=\frac1n\sum_{j=1}^{n}\frac{|\hat c_j-c_j|}{c_j+0.1}.\end{aligned}
\label{eq:metrics}
\end{equation}

\paragraph{Baselines.}
We compare with published results of RGB-based, skeleton-based, and LLM-based methods; skeleton-based methods count from poses estimated on the frames (Tab.~\ref{tab:reported_results}). For Fig.~\ref{fig:counting_comparison}, we run the public TransRAC checkpoint with 64 input frames, its original backbone, and start sampling.

\paragraph{Implementation details.}
The query readout uses rank-32 LoRA, up to eight 121-frame windows per video with a 60-frame hop, $H=D=96$, $J=32$ queries, and $B=64$ bins; affinities come from one denoising pass at noise strength 0.35. R1 is trained for up to 120 epochs (batch size 8, learning rate $3\times10^{-4}$) and R2 is fine-tuned from it for up to 20 epochs (learning rate $10^{-5}$), with checkpoints selected by validation OBO minus nMAE (Appendix~\ref{sec:arxiv_implementation}).

\subsection{Main Results}
\label{sec:benchmark_results}

\paragraph{Comparison with published results.}
DiffRAC achieves the best nMAE and OBO on all three benchmarks among the methods in Tab.~\ref{tab:reported_results}. On RepCount-A, it improves OBO by 18.0 points over ESCounts and by 10.4 points over CountLLM, and lowers nMAE from 0.162 to 0.143. The gain is not due to refinement alone: a single query readout already reaches 64.25\% mean OBO (Tab.~\ref{tab:repcount_ablation}). On UCFRep, DiffRAC counts 92 of the 105 videos within one repetition (\UCFOBO\%), exceeding CountLLM by 3.7 percentage points. Its nMAE is \UCFOfficialMAE. On Countix, it improves OBO over CountLLM by 3.5 points and nMAE from 0.215 to 0.166. The strongest skeleton-based method, MoCount~\cite{mocount}, which relies on an external pose estimator, trails DiffRAC by 9.4 OBO points on RepCount-A and 7.1 points on UCFRep. As modality and protocol differ, we regard these comparisons as indicative.

\par\addvspace{0.8\baselineskip}
\noindent\begin{minipage}{\columnwidth}
\captionsetup{type=table}
\caption{\textbf{Transfer from RepCount-A to UCFRep and Countix} without fine-tuning on the target. Baseline scores are from CountLLM~\cite{countllm}; the RepNet result on UCFRep is for the modified variant described in~\cite{repnet_eval}. OBO is in percent. \textbf{Bold} and \underline{underline} mark the best and second-best value in each column.}
\label{tab:cross_dataset}
\centering
\small
\setlength{\tabcolsep}{3.5pt}
\renewcommand{\arraystretch}{1.0}
\begin{tabular}{lrrrr}
\toprule
\textbf{Method} & \multicolumn{2}{c}{UCFRep} & \multicolumn{2}{c}{Countix} \\
\cmidrule(lr){2-3}\cmidrule(lr){4-5}
 & nMAE $\downarrow$ & OBO $\uparrow$ & nMAE $\downarrow$ & OBO $\uparrow$ \\
\midrule
RepNet~\cite{repnet} & 0.998 & 0.90 & 0.729 & 18.30 \\
Context~\cite{zhang2020context} & 0.762 & 41.20 & 0.654 & 20.80 \\
TransRAC~\cite{transrac} & 0.640 & 32.40 & 0.593 & 36.40 \\
MFL~\cite{li2024mfl} & 0.523 & 35.00 & -- & -- \\
ESCounts~\cite{escount} & 0.317 & 57.10 & 0.374 & 52.10 \\
CountLLM~\cite{countllm} & \textbf{0.139} & \underline{81.60} & \underline{0.281} & \underline{73.00} \\
\midrule
DiffRAC & \underline{0.181} & \textbf{82.43} & \textbf{0.265} & \textbf{73.87} \\
\bottomrule
\end{tabular}
\end{minipage}\par\addvspace{0.8\baselineskip}

\subsection{Cross-Dataset Generalization}
\label{sec:cross_dataset}

Trained on RepCount-A and evaluated without fine-tuning (Tab.~\ref{tab:cross_dataset}), DiffRAC attains the highest OBO on both targets, 82.43\% on UCFRep and 73.87\% on Countix, exceeding CountLLM by 0.83 and 0.87 points and ESCounts by 25.3 and 21.8 points. It also has the lowest Countix nMAE (0.265 \vs 0.281), whereas its UCFRep nMAE is higher than CountLLM's (0.181 \vs 0.139), suggesting fewer but larger misses.

\par\addvspace{0.8\baselineskip}
\noindent\begin{minipage}{\columnwidth}
\captionsetup{type=table}
\caption{\textbf{Paired ablations on the RepCount-A test split.} Means over three matched seeds with a common prompt; checkpoints are selected by validation OBO minus nMAE. OBO uses all 152 videos and nMAE the 151 positive-count videos. Panel~A uses the first readout; Panel~B uses dual-readout fusion.}
\label{tab:repcount_ablation}
\centering
\small
\setlength{\tabcolsep}{3.5pt}
\resizebox{\linewidth}{!}{%
\begin{tabular}{@{}lrr@{}}
\toprule
Configuration & OBO (\%) $\uparrow$ & nMAE $\downarrow$ \\
\midrule
\multicolumn{3}{@{}l}{\textit{A. Representation and period branch (first readout)}} \\
Hidden-feature self-similarity & 35.09 & 0.394 \\
Hidden-feature self-similarity + period branch & 38.82 & 0.314 \\
Q--K cross-frame relations & 57.68 & 0.203 \\
Q--K relations + period branch & 64.25 & 0.178 \\
\midrule
\multicolumn{3}{@{}l}{\textit{B. Adaptation and aggregation (dual-readout fusion)}} \\
Matched full configuration & 74.34 & 0.143 \\
Without LoRA & 57.02 & 0.199 \\
Without coverage normalization & 32.24 & 0.372 \\
\bottomrule
\end{tabular}
}
\end{minipage}\par\addvspace{0.8\baselineskip}

\begin{figure*}[!t]
\centering
\includegraphics[width=\textwidth]{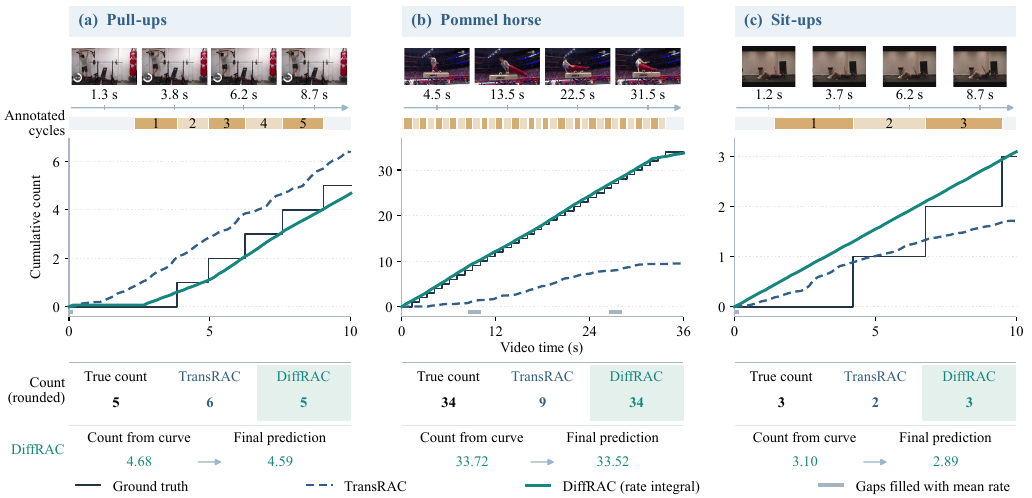}
\caption{\textbf{Cumulative counts on three RepCount-A test videos.} Black: annotated cycle ends; blue dashed: integrated TransRAC density; green: DiffRAC's coverage-normalized rate integral, with gray bins filled by its mean rate. Below each plot: rounded counts, and DiffRAC's rate integral and final prediction.}
\label{fig:counting_comparison}
\end{figure*}

\subsection{Qualitative Analysis}
\label{sec:qualitative_counting}

\paragraph{Period evidence and count on a test video.}
In the 14-second pull-up clip of Fig.~\ref{fig:mechanism_real}, the off-diagonal bands in the W4 affinity map show that the denoiser relates corresponding phases of successive pull-ups. All five period posteriors peak at 1.67\,s, comparable to the average cycle duration of about 1.75\,s (eight repetitions in 14\,s). The rates nevertheless vary from 0.495 to 0.585 repetitions per second, because activity and frequency also depend on the local node features. Coverage-normalized integration gives 7.65 repetitions, and calibration raises the estimate to 7.90, which rounds to the annotated eight.

\paragraph{Comparison with TransRAC.}
Figure~\ref{fig:counting_comparison} contrasts density regression with the rate formulation on three typical cases. (a)~Before the first annotated cycle ends, TransRAC's density already accumulates count, whereas DiffRAC's curve stays nearly flat. TransRAC predicts 6, while DiffRAC's rate integral reaches 4.68 and its calibrated prediction is 4.59, which rounds to the annotated count of 5. (b)~The pommel-horse clip contains 34 short, fast cycles. TransRAC undercounts severely (9), while DiffRAC's rate integral follows the annotated staircase and reaches 33.72. Its calibrated prediction of 33.52 rounds to 34. (c)~For three slow sit-ups, TransRAC undercounts (2). DiffRAC rises at a nearly constant rate, with a rate integral of 3.10 and a calibrated prediction of 2.89 that rounds to 3. Its curve does not reproduce the individual steps, however, because a window rate spreads each cycle over the window rather than localizing its end (Sec.~\ref{sec:discussion}).

\subsection{Ablation Studies}
\label{sec:repcount_ablation}

Each row of Tab.~\ref{tab:repcount_ablation} averages three matched seeds on the RepCount-A test split, with checkpoints selected on validation. Panel~A evaluates the first readout (R1); Panel~B uses dual-readout fusion (Sec.~\ref{sec:refinement}).

\paragraph{Diffusion Q--K affinities \vs hidden-feature similarity.}
The hidden-feature self-similarity baseline computes cosine similarity on the inputs of the same attention layers of the same adapted Wan, pooled and normalized identically, under the same readout. Q--K affinities raise OBO from 35.09\% to 57.68\% without the period branch and from 38.82\% to 64.25\% with it, and win for every seed and both readouts by 17.8 to 38.2 points. This supports the hypothesis of Sec.~\ref{sec:intro} that attention between corresponding phases is a better period cue than feature similarity.

\paragraph{Period evidence.}
The period branch raises Q--K OBO from 57.68\% to 64.25\% (Appendix~\ref{sec:arxiv_ablation}). The template scores are the only evidence computed on the learned pair features.

\paragraph{Adaptation and aggregation.}
Removing LoRA lowers OBO from 74.34\% to 57.02\%. Replacing coverage normalization with direct accumulation, at fixed weights and window predictions, lowers OBO to 32.24\% and raises nMAE from 0.143 to 0.372, because shared frames are counted repeatedly.

\paragraph{Rate integration and calibration.}
At fixed weights before calibration, the R1 rate integral already reaches 40.57\% OBO with Q--K affinities and period evidence, \vs 25.88\% without period evidence and 7.89\% for hidden-feature self-similarity with period evidence; calibration raises the Q--K readout to 64.25\%. The gains thus originate in the window rates rather than in calibration.

\subsection{Counting Error Analysis}
\label{sec:error_patterns}

\paragraph{Effect of count refinement.}
On Countix, teacher-local reranking mainly corrects counts of 5--19, where the teacher is often close to but outside the tolerance, while nMAE is essentially unchanged because small errors on counts below five carry a large normalized cost. Its $\pm3$ search range limits it at high counts: for most missed annotations with at least 20 repetitions, no candidate lies within one repetition of the annotation.
Camera-motion compensation with fixed initial predictions likewise helps mainly at counts of 5--9 and leaves the high-count group largely unchanged.

\section{Discussion and Limitations}
\label{sec:discussion}
\paragraph{Where errors remain.}
Temporal compression limits the shortest resolvable period, consistent with the undercounting of high-count Countix videos, and the 121-frame window can truncate slow cycles. Spatial pooling mixes the motion of several subjects, and misses also occur when cycle lengths vary little. The rate integral recovers totals but not cycle boundaries (Fig.~\ref{fig:counting_comparison}(c)) and is sensitive to long pauses: in one RepCount-A clip whose fifth cycle starts 525 frames after the fourth ends, DiffRAC predicts 11 for the annotated nine.

\paragraph{Computational cost.}
Each uncached window requires one forward pass of the 5B-parameter Wan; a one-minute 30-fps clip yields 29 windows before selection, and readout training reuses cached affinities.

\section{Conclusion}
We presented DiffRAC, which counts repetitions from the cross-frame Q--K affinities of a LoRA-adapted video diffusion transformer: it aggregates them into period evidence, decodes window rates as activity times frequency, and integrates the rates once on the original timeline. DiffRAC reaches 74.34\% OBO on RepCount-A and \UCFOBO\% on UCFRep, exceeds CountLLM's OBO when transferred to UCFRep and Countix, and matched ablations trace its gains to the Q--K representation, the period evidence, and coverage normalization.

The remaining errors point to future work. Finer temporal resolution, or adaptive window lengths, could address the undercounting of fast, high-count videos and the truncation of slow cycles. Replacing spatial pooling with subject-aware pooling would separate the motion of several people. A per-cycle readout that localizes cycle ends would turn the rate integral into an interpretable segmentation. Finally, distilling the diffusion affinities into a lighter encoder would reduce the cost of one diffusion forward pass per window.

\clearpage
\appendix
\raggedbottom
\setcounter{topnumber}{3}
\setcounter{dbltopnumber}{1}
\setcounter{figure}{0}
\setcounter{table}{0}
\setcounter{equation}{0}
\renewcommand{\thefigure}{S\arabic{figure}}
\renewcommand{\thetable}{S\arabic{table}}
\renewcommand{\theequation}{S\arabic{equation}}
\renewcommand{\theHfigure}{supp.\arabic{figure}}
\renewcommand{\theHtable}{supp.\arabic{table}}
\renewcommand{\theHequation}{supp.\arabic{equation}}
\twocolumn[{\centering\Large
\textbf{\thetitle}\par
\vspace{0.5em}
Appendices\par\vspace{1em}}]
The appendices provide implementation and training details for the period-aware query readout (Appendix~\ref{sec:arxiv_implementation}), evaluation protocols (Appendix~\ref{sec:arxiv_evaluation}), and paired ablations on RepCount-A (Appendix~\ref{sec:arxiv_ablation}).

\section{Implementation and Training Details}
\label{sec:arxiv_implementation}

\begin{figure*}[t]
\centering
\includegraphics[width=\textwidth]{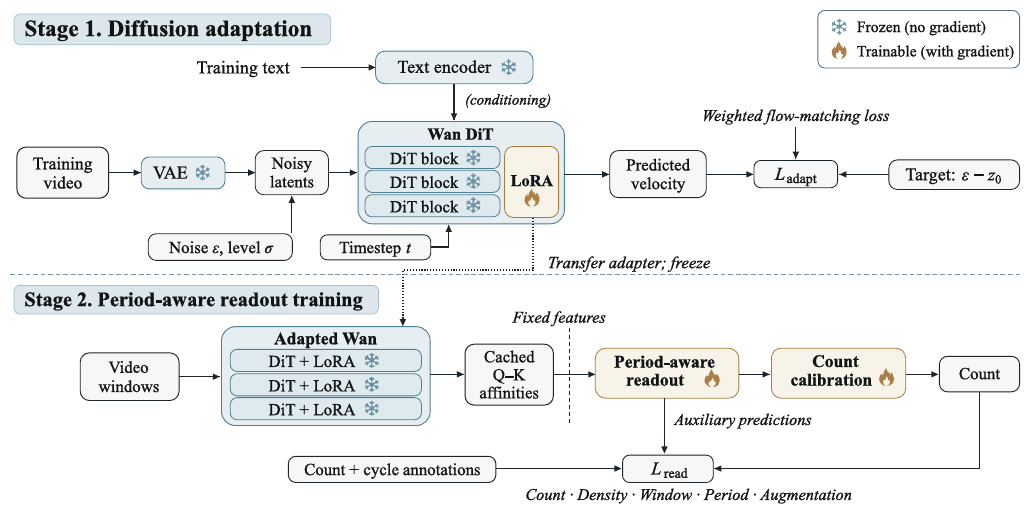}
\caption{\textbf{Two-stage training of DiffRAC.} Stage 1 adapts Wan through LoRA using the flow-matching objective $\mathcal{L}_{\mathrm{adapt}}$, while keeping the pretrained DiT, VAE, and text encoder frozen. Stage 2 freezes the adapted encoder, including its LoRA parameters, and caches cross-frame Q--K affinities. The period-aware readout and count calibrator are then optimized on these fixed features with $\mathcal{L}_{\mathrm{read}}$, using count and cycle annotations as supervision. Snowflakes indicate frozen modules, and flames indicate trainable modules.}
\label{fig:arxiv_training_stages}
\end{figure*}

\paragraph{Training overview.} DiffRAC separates diffusion adaptation from readout learning (Fig.~\ref{fig:arxiv_training_stages}). We first adapt Wan~\cite{wan,wan22} to counting videos using LoRA~\cite{lora} and the flow-matching objective~\cite{fm}, with balanced intact windows and temporally shuffled negatives. The adapters are optimized with AdamW at a learning rate of $10^{-4}$. When the denoiser is conditioned on the first frame, its latent node is excluded from the loss.

We then freeze the adapted backbone and its LoRA parameters and cache Q--K affinities for all training windows. On these fixed features, we optimize the period-aware query readout and count calibrator using count and cycle annotations under the objective in Eq.~\eqref{eq:loss}. Appendix~\ref{sec:arxiv_readout_training} specifies the supervision targets and shuffled-affinity augmentation.

\subsection{Temporal Affinity Extraction}
\label{sec:arxiv_extraction}
\paragraph{Inference prompt.} Affinity extraction for the two RepCount-A query readouts uses the fixed template: \emph{Repetitive actions are characterized by periodic temporal intervals, consistent motion patterns, and visual similarity across cycles. The repeated action is \{action\}. Count complete repetitions in the video.} Only \{action\} varies across videos and is supplied by the dataset. The prompt contains no reference count or annotation-derived window answer.

\paragraph{Window sampling.} The query configuration uses 121-frame windows with candidate starts spaced 60 original frames apart and a final window aligned to the video endpoint. If more than eight candidates are available, we retain eight at evenly spaced candidate indices, rounded to the nearest integer, including both endpoints. Valid original-frame indices are retained. Attention softmax covers all video keys before spatial averaging. Other extraction settings follow Secs.~\ref{sec:representation} and~\ref{sec:configs}.

\paragraph{Computational cost.} For a video of $N$ frames, window span $L$ and hop $h$, endpoint alignment yields
\[
W_{\mathrm{cand}}=1+\max\!\left(0,\left\lceil\frac{N-L}{h}\right\rceil\right)
\]
candidate windows. With $L=121$ and $h=60$, a 60-second video at 30 fps yields 29 candidates, of which the query configuration retains at most eight. Each retained window requires one diffusion-backbone forward pass to compute its affinities. Subsequent readout training reuses the cached affinities without additional backbone passes.

\subsection{Period Evidence and Window Rate Decoding}
\label{sec:arxiv_period}
The lag encoder uses two one-dimensional convolutions of kernel width 3 and dilations 1 and 2, followed by averaging over lags. The posterior covers all 29 candidate periods, including $p=1$, whose centered spectral term is zero.

For period $p$, the template measures each node separation's distance to the nearest period multiple:
\begin{equation}
\begin{aligned}
d_{ij}&=|i-j|,\qquad r_{p,ij}=d_{ij}\bmod p,\\
e_{p,ij}&=\min(r_{p,ij},p-r_{p,ij}),\\
H_p[i,j]&=\mathbf1[i\ne j]\exp\!\left(-\frac{e_{p,ij}^2}{2\sigma_p^2}\right),\\
\sigma_p&=\max(p/4,1),\qquad G_p=\frac{H_p}{\|H_p\|_F+\epsilon}.
\end{aligned}
\label{eq:arxiv_period_template}
\end{equation}
We use $\epsilon=10^{-6}$. For the learned pair-score map $U\in\mathbb R^{T\times T}$, the projection $\psi[p]=\sum_{i,j}U[i,j]G_p[i,j]$ supplies template evidence to the period head.

A $1\times1$ projection and three residual blocks encode the 96 affinity channels at width 96. Each block contains two $3\times3$ convolutions with GroupNorm and GELU. Two cross-attention blocks implement the 32-query decoder of Sec.~\ref{sec:decoding}. The decoder combines local node features with period evidence to estimate repetition activity and frequency, whose product defines the window counting rate.

\subsection{Coverage-Normalized Rate Integration}
\label{sec:arxiv_integration}
\begin{figure*}[t]
\centering
\includegraphics[width=0.85\textwidth]{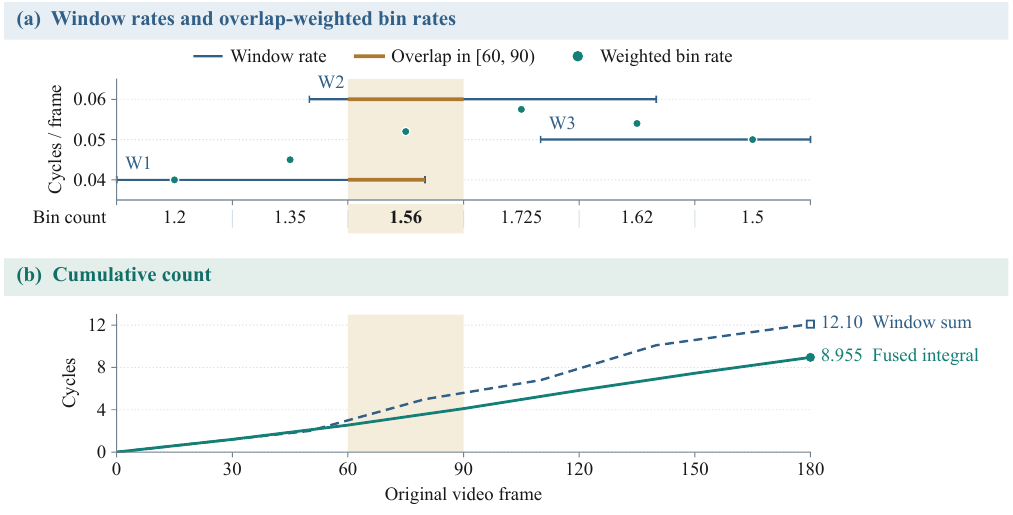}
\caption{\textbf{Coverage-normalized rate integration on a constructed example.} (a) Window rates, overlap-weighted bin rates and the resulting bin counts. The shaded interval marks $[60,90)$. Six bins illustrate the 64-bin grid used by DiffRAC. (b) The fused rate integral gives 8.955 cycles, compared with 12.10 from directly summing window counts. Both quantities precede boundary correction and calibration.}
\label{fig:arxiv_coverage_example}
\vspace{-8pt}
\end{figure*}

We integrate window rates on the original timeline using the overlap weighting and uncovered-bin filling of Sec.~\ref{sec:aggregation}. Figure~\ref{fig:arxiv_coverage_example} illustrates this operation with three constant-rate windows. In the bin $[60,90)$, two windows cover 20 and 30 frames at rates of 0.04 and 0.06 cycles per frame. The overlap-weighted rate is $(20\times0.04+30\times0.06)/(20+30)=0.052$, giving a bin count of $30\times0.052=1.56$. This averaging prevents overlapping windows from contributing their full counts independently.

\subsection{Readout Training}
\label{sec:arxiv_readout_training}
Within Stage 2, R1 is the initially trained query readout, and R2 is further fine-tuned from its checkpoint. Both operate on frozen diffusion affinities. Table~\ref{tab:arxiv_readout_config} specifies their configurations. Query checkpoints maximize validation OBO minus nMAE, with ties resolved by higher OBO.

\begin{table}[!htbp]
\caption{\textbf{Query-readout configurations.} R1 and R2 denote the initial and fine-tuned query readouts, respectively. Epoch indices start at zero. R1 uses a batch size of 8. Both runs use AdamW with weight decay $10^{-4}$ and gradient clipping at norm 1.}
\label{tab:arxiv_readout_config}
\centering\small
\setlength{\tabcolsep}{6pt}
\begin{tabular}{@{}lrr@{}}
\toprule
Setting & R1 & R2 \\
\midrule
Input channels & 96 & 96 \\
Width & 96 & 96 \\
Inference window cap & 8 & 8 \\
Learning rate & $3\!\times\!10^{-4}$ & $10^{-5}$ \\
Epoch budget & 120 & 20 \\
Selected epoch & 88 & 1 \\
Seed & 2027 & 2056 \\
\bottomrule
\end{tabular}
\end{table}

\paragraph{Supervision targets.} Each annotated cycle contributes a unit-mass Gaussian centered at its midpoint~\cite{transrac}. Its standard deviation is one sixth of the cycle span, and we integrate it over bins within that span. Single-bin cycles receive unit mass directly. For a video of reference count $y>0$, each valid window's period target is a discrete Gaussian with standard deviation 1.5 centered at $N/(y s_w)$, where $s_w$ denotes original frames per latent step. Integrating the density over $O_w$ gives $y_w$, with rate target $y_w/|O_w|$ and activity target $\mathbf1[y_w>0.05]$. Joint row--column permutations of the affinities receive zero count and density targets. Table~\ref{tab:arxiv_readout_weights} lists the objective coefficients.

\begin{table}[!htbp]
\caption{\textbf{Query-readout objective coefficients.} The augmentation weight scales the additional readout objective evaluated on shuffled inputs.}
\label{tab:arxiv_readout_weights}
\centering\small
\setlength{\tabcolsep}{9pt}
\begin{tabular}{@{}lrr@{}}
\toprule
Term & R1 & R2\\
\midrule
Normalized count & 1 & 1.2\\
Absolute count & 0.02 & 0.02\\
Log count & 0.30 & 0.70\\
Normalized density & 0.05 & 0.10\\
Raw density & 0.02 & 0.03\\
Window count & 0.20 & 0.20\\
Window rate & 0.10 & 0.20\\
Window activity & 0.05 & 0.05\\
Period KL & 0.02 & 0.02\\
Period entropy & 0.001 & 0.001\\
Shuffled-zero augmentation & 0.02 & 0.02\\
\bottomrule
\end{tabular}
\end{table}

\section{Evaluation Protocols}
\label{sec:arxiv_evaluation}
\paragraph{Data and metrics.} Dataset splits and metric definitions follow Sec.~\ref{sec:configs} and Eq.~\eqref{eq:metrics}. UCFRep reference counts are derived from cycle boundaries. RepCount-A ablation averaging is specified in Appendix~\ref{sec:arxiv_ablation_protocol}.

\paragraph{Countix annotation coverage.} Filtering the ESCounts test list~\cite{escount} for positive counts and more than 40 frames gives 2,270 annotations from 2,258 clips. Evaluation uses 2,262 available predictions from 2,254 clips. Four clips, each with two annotations, lack predictions. Repeated annotations are evaluated as separate rows.

\section{Additional Ablation Studies}
\label{sec:arxiv_ablation}
\subsection{Experimental Protocol}
\label{sec:arxiv_ablation_protocol}
We follow the training budgets and checkpoint criterion in Appendix~\ref{sec:arxiv_readout_training}, using seed pairs 2027/2056, 2028/2057 and 2029/2058. OBO uses all 152 RepCount-A test videos and nMAE the 151 positive-count videos. Individual readouts are evaluated before external gating, fusion and the count offset.

The representation control computes cosine similarity from spatially pooled channel groups of the same Wan layers' self-attention inputs, followed by the Q--K temporal normalization. Removing the period branch disables its prediction, auxiliary losses, node conditioning, lag/period readout inputs and calibration statistics.

\subsection{Representation and Period Evidence}
Q--K outperforms hidden-feature self-similarity in every matched run of both readouts (Table~\ref{tab:arxiv_rep_paired}). The period branch improves R2 performance, although its benefit is not uniform across R1 Q--K seeds.

\begin{table}[!htbp]
\caption{\textbf{Variation across paired seeds.} Minimum--maximum OBO differences across three matched seeds, in percentage points. Self denotes hidden-feature self-similarity and P the period branch. Removing P reports without minus with P.}
\label{tab:arxiv_rep_paired}
\centering\small
\setlength{\tabcolsep}{2pt}
\begin{tabular}{@{}lrr@{}}
\toprule
Contrast & R1 & R2 \\
\midrule
Q--K minus Self, P & $[+17.76,+30.26]$ & $[+18.42,+25.00]$ \\
Q--K minus Self, no P & $[+18.42,+25.66]$ & $[+20.39,+38.16]$ \\
Remove P from Q--K & $[-13.16,+3.95]$ & $[-8.55,-2.63]$ \\
Remove P from Self & $[-6.58,-0.66]$ & $[-19.08,-4.61]$ \\
\bottomrule
\end{tabular}
\end{table}

\subsection{Rate Integration and Calibration}
Table~\ref{tab:arxiv_r1_aggregation} compares the R1 aggregation stages at fixed trained weights, following Sec.~\ref{sec:repcount_ablation}. The advantage of Q--K affinities and period evidence is already apparent in the uncalibrated rate integral. Calibration further improves counting accuracy while preserving the ranking of the three configurations.

\begin{table}[!htbp]
\caption{\textbf{R1 aggregation stages.} Mean OBO (\%) over three matched seeds on all 152 RepCount-A test videos. We hold each trained readout fixed and compare the rate integral with the output after boundary correction and calibration, before external fusion and the fixed count offset. Self denotes hidden-feature self-similarity and P the period branch.}
\label{tab:arxiv_r1_aggregation}
\centering\small
\setlength{\tabcolsep}{4pt}
\begin{tabular}{@{}lrr@{}}
\toprule
Input & Integral $\uparrow$ & + Calibration $\uparrow$ \\
\midrule
Self + P & 7.89 & 38.82 \\
Q--K & 25.88 & 57.68 \\
Q--K + P & \textbf{40.57} & \textbf{64.25} \\
\bottomrule
\end{tabular}
\end{table}

Related studies investigate task-specific diffusion representations
through multimodal alignment~\cite{mmrepa} and image-restoration
fidelity analysis~\cite{areir}.
Memory management supports long-video understanding and
generation~\cite{jin2026videomem,zhu2025memorize}, while adaptive
reasoning controls the computational cost of video
understanding~\cite{tian2026adathinkv,tian2026delayed}.

\end{document}